\documentclass[a4paper,11pt,fleqn]{article}
\usepackage{graphicx}
\usepackage{amsmath,amssymb,xfrac,bbm,dsfont}
\usepackage[utf8]{inputenc}

\usepackage{pstricks,dsfont}
\usepackage{amsmath}
\usepackage{amssymb}
\usepackage{mathrsfs}
\usepackage{theorem} 
\usepackage{booktabs}
\usepackage{euscript}
\usepackage{exscale,relsize}
\usepackage{graphicx}
\usepackage{booktabs}
\usepackage{srcltx}
\usepackage{xcolor}
\usepackage{charter}
\usepackage{url}

\allowdisplaybreaks

\newtheorem{theorem}{Theorem}[section]

\newtheorem{proposition}[theorem]{Proposition}

\theoremstyle{plain}{\theorembodyfont{\rmfamily}%
}
\theoremstyle{plain}{\theorembodyfont{\rmfamily}%
}
\theoremstyle{plain}{\theorembodyfont{\rmfamily}%
\newtheorem{remark}[theorem]{Remark}}
\theoremstyle{plain}{\theorembodyfont{\rmfamily}%
}
\theoremstyle{plain}{\theorembodyfont{\rmfamily}%
}
\theoremstyle{plain}{\theorembodyfont{\rmfamily}%
\newtheorem{definition}[theorem]{Definition}}
\theoremstyle{plain}{\theorembodyfont{\rmfamily}
}
\theoremstyle{plain}{\theorembodyfont{\rmfamily}
}

\numberwithin{equation}{section}
\begin{document}

\title{\sffamily\LARGE General Fair Empirical Risk Minimization\thanks{
This work was supported in part by both SAP SE and Amazon Web Services.}}

\author{Luca Oneto \\
\small DIBRIS, University of Genova (Italy) \\
\small {luca.oneto@unige.it} \\[3mm] 
Michele Donini \\
\small  Amazon Web Services (US) \\
\small {donini@amazon.com}\\[3mm]
Massimiliano Pontil \\
\small  CSML, Istituto Italiano di Tecnologia (Italy) \\
\small {massimiliano.pontil@iit.it} \\
\small and \\
\small University College London (UK)
}

\maketitle
\begin{abstract}
We tackle the problem of algorithmic fairness, where the goal is to avoid the unfairly influence of sensitive information, in the general context of regression with possible continuous sensitive attributes.
We extend the framework of fair empirical risk minimization of~\cite{OnetoC060} to this general scenario, covering in this way the whole standard supervised learning setting. 
Our generalized fairness measure reduces to well known notions of fairness available in literature.
We derive learning guarantees for our method, that imply in particular its statistical consistency, both in terms of the risk and the fairness measure.
We then specialize our approach to kernel methods and propose a convex fair estimator in that setting.
We test the estimator on a commonly used benchmark dataset (Communities and Crime) and on a new dataset collected at the University of Genoa\footnote{The data and the research are related to the project DROP@UNIGE of the University of Genoa.}, 
containing the information of the academic career of five thousand students.
The latter dataset provides a challenging real case scenario of unfair behaviour of standard regression methods that benefits from our methodology.
The experimental results show that our estimator is effective at mitigating the trade-off between accuracy and fairness requirements.
\end{abstract}
\section{Introduction}
The problem of designing learning methods that do not use sensitive information in a discriminatory way (e.g.~knowledge about the ethnic group of an individual, sex, age) is receiving increasing attention, due to its fundamental importance in real-life scenarios, see e.g.~\cite{pleiss2017fairness,beutel2017data,hardt2016equality,feldman2015certifying,woodworth2017learning,zafar2017fairness,zafar2017parity,zafar2017fairnessARXIV,kamishima2011fairness,kearns2017preventing,perez2017fair,berk2017convex,adebayo2016iterative,calmon2017optimized,kamiran2009classifying,zemel2013learning,kamiran2012data,kamiran2010classification} and references therein.
In this paper we follow a recent line of work~\cite{OnetoC060,agarwal2017reductions,zafar2017fairness,menon2018cost,bechavod2018Penalizing,zafar2017fairnessARXIV,kamishima2011fairness,kearns2017preventing,perez2017fair,berk2017convex,alabi2018optimizing,dwork2018decoupled} in which the fairness constraint is directly taken into account during the learning procedure.
An important departure from previous work that we take in this paper is to consider the possibility that the sensitive feature and/or the output (response variable) we wish to predict take real values.

% In several real-world scenarios, classification methods are not sufficient to solve the task at hand.
The importance of being able to solve regression tasks and possibly dealing with continuous sensitive features can be highlighted by the following example.
At the University of Genoa, automatic systems are needed to predict students' performance for the purpose of improving the teaching quality and the students' support systems.
In this case, the response variable is the course mark and the sensitive features can be both categorical (e.g.~sex or ethnic group) or continuous (e.g.~age or financial status).

Common notions of fairness that have been used in the setting of classification with categorical sensitive features is that of Equal Opportunity or Equalized Edds~\cite{hardt2016equality}.
They aim to balance decisions of a classifier among the different sensitive groups and label sets.
We show how these notions can be extended to the general supervised learning setting (regression and classification) with general sensitive features (categorical and continuous).
We observe that these novel fairness constraints can be incorporated within the Empirical Risk Minimization (ERM) framework.
Our method and analysis build up and extend the Fair ERM (FERM) framework developed in~\cite{OnetoC060}.
As the fairness measures used here are more general than those employed in that work, we name our approach General FERM (G-FERM).
We show that G-FERM is supported by consistency guarantees both in terms of risk and fairness measure.
Specifically, we derive both risk and fairness bounds, which support the statistically consistency of G-FERM.
We give a concrete instance of G-FERM in the setting of kernel methods, leading to a form of constrained regularized empirical risk minimization, in which the fairness constraint is obtained by composting the $\ell_1$ norm with a linear transformation.

{\bf Contributions.}
First, we present new generalized notions of fairness that encompass well studied notions used for classification and regression with categorical and numerical sensitive feature.
Second, we study statistical bounds for G-FERM that imply consistency properties both in terms of fairness measure and risk of the selected model.
As a third contribution, we instantiate G-FERM in the setting of kernel methods, leading to an efficient convex estimator.
We test this estimator on a commonly used benchmark dataset (Communities and Crime) and on a new dataset collected at University of Genoa, containing the information of the academic career of five thousand students.
The latter dataset provides a challenging real case scenario of unfair behaviour of standard methods for regression that is solvable by using our methodology.
The experimental results show that our estimator is effective at mitigating the trade-off between accuracy and fairness requirements.

{\bf Paper Organization.}
In Section~\ref{sec:2} we discuss previous work on fairness, with a particular focus on regression and or continuous sensitive features.
In Section~\ref{sec:3} we introduce our notion of fairness which leads us to the G-FERM and study its statistical properties.
In Section~\ref{sec:4} we give the kernel-based G-FERM estimator and in Section~\ref{sec:5} report on numerical experiments on two real datasets.
Finally in Section~\ref{sec:6} we draws conclusions and comment on future research directions.
\section{Related works}
\label{sec:2}
In the context of fairness, most of the papers in literature address the problem of binary classification task with categorical (or even binary) sensitive features~\cite{hardt2016equality,zafar2017fairness}; a broad review on classification with categorical sensitive feature is provided in~\cite{OnetoC060}.
This task is indeed very important, because it is strictly related to the possibility of having access to specific benefits (e.g.~loans) without being discriminated due to gender or ethnic characteristics.
On the other hand, the set of problems solvable by using these methods is limited and not comprehensive of all the real-world case scenarios.

Focusing on the works able to handle regression tasks, we can divide them by the type of problems they are able to solve and the notion of fairness they exploit.
As we will see, with very few exceptions -- e.g.~\cite{komiyama2017two} -- most of the methods in literature are not able to deal with both classification and regression task and with both numerical and categorical sensitive features with an unified approach supported by theoretical consistency results.
In fact, they introduce task oriented notions of fairness and/or do address the statistical consistency of their method with respect to the risk and the fairness measure employed.
%they fail in presenting 
%consistency results with respect to the risk and fairness of the optimization problem they solve.

The largest family of methods tackle regression problems with (single) categorical or binary sensitive feature~\cite{berk2017convex,calders2013controlling,fitzsimons2018equality,raff2017fair}.
For example, in~\cite{berk2017convex}, a convex approach for regression is proposed, where the authors use a specific definition of fairness in order to have models which treat similar examples in a similar way, in the sense of the predicted outcome.
The authors tackle the problem by introducing a new convex regularizer and by imposing this notion on different regression tasks.
Another example is~\cite{fitzsimons2018equality}, where the authors use an adapted version of Demographic Parity~\cite{dwork2012fairness} for classification, in the context of regression.

Reducing the regression problem to have only categorical sensitive features is a serious limitation.
In this sense, few interesting papers present regression methods able to deal with continuous sensitive attributes~\cite{komiyama2017two,komiyama2018nonconvex,perez2017fair}.
Differently to our approach, the authors impose other definitions of fairness (e.g.~Disparate Impact~\cite{zafar2017fairness} or even ad-hoc brand new definitions).
Moreover, it is important to note that these methods do not naturally extend to the case of not-continuous sensitive attributes. 

Considering a larger spectrum of possible methodologies, it is possible to find in literature other methods able to solve regression tasks by imposing some concept of fairness.
\cite{nabi2018fair} and~\cite{nabi2018learning} tackle the regression problem exploiting the causal machine learning framework.
These methods can handle potentially both continuous and categorical sensitive features.
The authors' analysis considers only the case of categorical ones, leaving the evolution to continuous sensitive attributes as possible future works.
Another interesting idea, presented in~\cite{yona2018probably}, is to study the fairness as a property of the metric of the feature space.
The authors introduce a new definition of metric-related fairness allowing them to solve a regression problem with categorical and continuous sensitive attributes.
Finally, learning fair pre-processing rules is another possible way to obtain a regression model that is fair.
In fact, for example in~\cite{zemel2013learning}, the fair representation of the data can be used in synergy with any classic regression method, in order to generate a fair regression model.
\section{Learning with Fairness Constraints}
\label{sec:3}
In this section, we introduce our framework for learning under fairness constraints.
We first recall some notation used throughout this work in Section~\ref{sec:3.1}.
We then present the proposed fairness measures in Section~\ref{sec:3.2}, which lead us to consider in Section~\ref{sec:3.3} a generalized version of the FERM approach~\cite{OnetoC060}.
Finally in Section~\ref{sec:3.4} we discuss the statistical properties of our method.
\subsection{Setting}
\label{sec:3.1}
Let $\mathcal{D} {{=}} \{ (\boldsymbol{x}_1,s_1,y_1),$ $\dots,$ $(\boldsymbol{x}_n,s_n,y_n) \}$ be a training set formed by $n$ samples drawn independently from an unknown probability distribution $\mu$ over $\mathcal{X} {{\times}} \mathcal{S} {{\times}} \mathcal{Y}$, where 
$\mathcal{X}$ is the input space, $\mathcal{S}$ is the space of the sensitive attribute and $\mathcal{Y}$ is the output space.
Both $\mathcal{S}$ and $\mathcal{Y}$ may be finite or continuous; if $\mathcal{Y}$ is a finite set of labels we are dealing with the classification setting and if $\mathcal{Y} \subseteq \mathbb{R}$ we are dealing with the regression setting.

Let $K$ and $Q$ be positive integers and define the sets:
\begin{equation*}
    \mathcal{Y}_K {=} \{ t_1, {\cdots}, t_{K+1} \} {\subset} \mathbb{R}^{K+1} \,\,\,\, \text{  and  } \,\,\,\, \mathcal{S}_Q {=} \{ \sigma_1, {\cdots}, \sigma_{Q+1} \} {\subset} \mathbb{R}^{Q+1},
\end{equation*} where $t_1 {<} t_2 {<} {\cdots} {<} t_{K+1}$, and $\sigma_1 {<} \sigma_2 {<} {\cdots} {<} \sigma_{Q+1}$.
The sets $\mathcal{Y}_K$ and $\mathcal{S}_{Q}$ are prescribed by the user:
the discretization process is driven by the application at hand and points in the same interval are regarded as indistinguishable.
For example, it does not make sense to state that a group of students at the University of Genoa is mistreated because the average grades are distant by less than $5\%$ of the mark range.
We also define, for every $1{\leq} k {\leq} K$ and $1{\leq} q {\leq} Q$, the subsets of training points $\mathcal{D}_{k,q} {{=}} \{(\boldsymbol{x}_i,s_i,y_i) : 1{\leq}i{\leq}n,
%(\boldsymbol{x},s,y) {{\in}} \mathcal{D}, 
y {\in} [t_k, t_{k{+}1}), s {\in} [\sigma_q, \sigma_{q{+}1}) \}$ and let $n_{k,q} {{=}} | \mathcal{D}_{k,q} |$.

We consider a function (or model) $f$ chosen from a set $\mathcal{F}$ of possible ones.
The functional form of the model may explicitly depend on the sensitive feature (i.e.~$f{:} \mathcal{X} {{\times}} \mathcal{S} {{\rightarrow}} \mathbb{R}$) or not (i.e.~$f{:} \mathcal{X} {{\rightarrow}} \mathbb{R}$) based on specific legal requirements in the application at hand~\cite{dwork2018decoupled,OnetoC062}.
For this reason we will indicate $f{:} \mathcal{Z} {\rightarrow} \mathbb{R}$ where $\mathcal{Z}$ may contain the sensitive feature (i.e.~$\mathcal{Z} {=} \mathcal{X} {{\times}} \mathcal{S}$) or not (i.e.~$\mathcal{Z} {=} \mathcal{X}$).
The error (risk) of $f$ is measured by a prescribed loss function $\ell{:}\mathbb{R} {{\times}} \mathcal{Y} {{\rightarrow}} \mathbb{R}$ .
The risk of a model $L(f)$, together with its empirical counterpart $\hat{L}(f)$, are defined respectively as
\begin{align}
L(f) {{=}} \mathbb{E} \left[ \ell(f(\boldsymbol{z}),y) \right], \nonumber 
\end{align}
and
\begin{align}
\textstyle
\hat{L}(f) {{=}} \frac{1}{n} \sum_{(\boldsymbol{z},y) {\in} \mathcal{D}} \ell(f(\boldsymbol{z}),y).
\nonumber
\end{align}
When necessary we will indicate with a subscript the particular loss function used and the associated risk, i.e.~$L_p(f) = \mathbb{E} \left[ \ell_p(f(\boldsymbol{z}),y) \right]$.

The purpose of a learning procedure is to find a model that minimizes the risk.
Since the probability measure $\mu$ is usually unknown, the risk cannot be computed, however we can compute the empirical risk and a natural learning strategy, called Empirical Risk Minimization (ERM), is then to minimize the empirical risk within a prescribed set of functions, see e.g.~\cite{shalev2014understanding}.
\subsection{$\epsilon$-Loss General Fair}
\label{sec:3.2}
In the literature different definitions of fairness of a classifier or real-valued function exist as described in Section~\ref{sec:2}.
It is important to stress that there is not yet a consensus about which definition 
should be employed to evaluate algorithmic fairness.
Moreover, most of the current fairness definitions are not able to deal with regression problems (or with continuous sensitive attributes), losing their meaning or being even not definable.
In this work we proposes a general notion of fairness able to deal with both classification and regression and with both categorical and numerical sensitive features and which generalizes previously known notions of fairness.
\begin{definition}
\label{def:fairnessGF}
A model $f$ is $\epsilon${-}general fair ($\epsilon${-}GF) with $\epsilon \in [0,1]$ if satisfies the following condition
\begin{align*}
\textstyle
\frac{1}{K Q^2}
\sum_{k {=} 1}^{K} \sum_{p,q {=} 1}^{Q} \left| P^{k,p}(f) - 
P^{k,q}(f)
\right| \leq {\epsilon}
\end{align*}
where, for every $1{\leq} k {\leq} K$ and $1{\leq} q{\leq} Q$, we have defined the conditional probabilities
\begin{align*}
P^{k,q}(f) {=} \mathbb{P}\! \left\{ \! f(\boldsymbol{z}) {\in} [t_k, t_{k{+}1}) \Big| y {\in} [t_k, t_{k{+}1}), s {\in} [\sigma_q, \sigma_{q{+}1})  \! \right\}  \!.
\end{align*}
\end{definition}
This definition says that a model is fair if its predictions are equally distributed independently of the value of the sensitive attribute.
It can be further generalized as follows.
\begin{definition}
\label{def:fairnessLGF}
For every $1{\leq} k {\leq} K$ let $\ell_k$ be a loss function.
For every $1{\leq} k {\leq} K$, $1{\leq} q {\leq} Q$, define 
the conditional risks
\begin{align*}
L^{k,q}_k(f) {=} \mathbb{E} \big[ \ell_k(f(\boldsymbol{z}),y) | y {\in} [t_k, t_{k{+}1}), s {\in} [\sigma_q, \sigma_{q{+}1}) \big].
\end{align*}
We say that a function $f$ is $\epsilon$-loss general fair ($\epsilon${-}LGF) with $\epsilon \in [0,1]$ if it satisfies the following condition
\begin{align*}
\textstyle
\frac{1}{K Q^2}
\sum_{k {=} 1}^{K} \sum_{p,q {=} 1}^Q \left| L^{k,p}_k(f) - L^{k,q}_k(f) \right| {\leq} \epsilon.
\end{align*}
\end{definition}
This definition says that a model is fair if its errors, relative to the loss function,
are approximately equally distributed independently of the value of the sensitive attribute.
Definition~\ref{def:fairnessLGF} includes Definition~\ref{def:fairnessGF} when we choose $ \ell_k({\hat y},y)$ ${=}$ $\mathds{1}\{ {\hat y} {\not\in} [t_k, t_{k+1}) \}$, for $1 {\leq} k {\leq}K$.
Moreover, it is possible to link Definition~\ref{def:fairnessLGF} to other fairness measures used before in the literature.
\begin{remark}
If we choose $\epsilon$ ${=}$ $0$, $\mathcal{Y}$ ${=}$ $ \{-1, +1 \}$, $\mathcal{S}$ ${=}$ $\{0, 1 \}$, $\mathcal{Y}_K$ ${=}$ $\{-1.5,$ $0,$ $ +1.5 \}$, $\mathcal{S}_Q$ ${=}$ $\{-0.5,$ $ 0.5,$ $ 1.5 \}$ and, for every $1{\leq}k{\leq}K$, let $\ell_k$ be the 0-1-loss, that is $\ell_k(y,y) {=} \mathds{1}\{y {\hat y} {\leq} 0\}$, then Definition~\ref{def:fairnessLGF} reduces to the notion of Equalized Odds~\cite{hardt2016equality,OnetoC060}.
On the other hand, in the same setting, if we let, for every $k$, $\ell_k$ be the linear loss, $\ell_k({\hat y},y) = (1 - y {\hat y})/2$, then we recover other notions of fairness introduced in~\cite{dwork2018decoupled}.
When $\epsilon {=} 0$, $\mathcal{Y} {\subseteq} \mathbb{R}$, $\mathcal{S} {=} \{0, 1 \}$, $\mathcal{Y}_K {=} \{-\infty, \infty \}$, $\mathcal{S}_Q {=}$ $ \{-0.5,$ $0.5,$ $1.5 \}$ then Definition~\ref{def:fairnessLGF} reduces to the notion of Mean
Distance introduced in~\cite{calders2013controlling} and also exploited in~\cite{komiyama2017two}.
Finally, in the same setting, if $\mathcal{S} {\subseteq} \mathbb{R}$ in~\cite{komiyama2017two} it is proposed to use the correlation coefficient which is equivalent to setting $\mathcal{S}_Q {=} \mathcal{S}$ in Definition~\ref{def:fairnessLGF}.
\end{remark}
\subsection{General Fair Empirical Risk Minimization}
\label{sec:3.3}
In this paper, we aim at minimizing the risk subject to a fairness constraint.
Specifically, we consider the problem
\begin{align}
\textstyle
\min_{f {\in} \mathcal{F}}\! \left\{ \! L(f) {:} 
\sum_{k {=} 1}^{K} \sum_{p,q {=} 1}^{Q} \left| L^{k,p}_k(f) {-} L^{k,q}_k(f) \right| {\leq} \epsilon
\right\}
\label{eq:alg:deterministic},
\end{align}
where $\epsilon \in [0,1]$ is the amount of unfairness that we are willing to bear.
Since the measure $\mu$ is unknown we replace the deterministic quantities with their empirical counterparts.
That is, we replace Problem~\eqref{eq:alg:deterministic} with
\begin{align}
\textstyle
\min_{f {\in} \mathcal{F}}\! \left\{ \! \hat{L}(f) {:}
\sum_{k {=} 1}^{K} \sum_{p,q {=} 1}^{Q} \left| \hat{L}^{k,p}_k(f) {-} \hat{L}^{k,q}_k(f) \right| {\leq} {\hat \epsilon}
\right\}
\label{eq:alg:empirical},
\end{align}
where $\hat{\epsilon} \in [0,1]$, and, for every $k {\in} \{1, {\cdots}, K \}$ and every $q {\in} \{1, {\cdots}, Q \}$ we defined the empirical conditional risks
\begin{align*}
\textstyle
\hat{L}^{k,q}_k(f) = \frac{1}{n_{k,q}} \sum_{(\boldsymbol{z},y) {\in} \mathcal{D}_{k,q}} \ell_k(f(\boldsymbol{z}),y).
\end{align*}
We will refer to~Problem~\eqref{eq:alg:empirical} as G{-}FERM since it generalizes the FERM approach introduced in~\cite{OnetoC060}.
\subsection{Statistical Analysis}
\label{sec:3.4}
Let $f^*$ be a solution of Problem~\eqref{eq:alg:deterministic}, and let $\hat{f}$ a solution of Problem~\eqref{eq:alg:empirical}.
In this section we will show that these solutions are linked one to another.
In particular, if the parameter $\hat{\epsilon}$ is chosen appropriately, we will show that, in a certain sense, the estimator $\hat{f}$ is consistent.
Our analysis extends the reasoning in~\cite{OnetoC060} to the more general setting presented here.

For this purpose, we require that for any data distribution, it holds with probability at least $1-\delta$ with respect to the draw of a dataset that
\begin{align}
\textstyle
\sup_{f \in \mathcal{F}} \big|L(f) - \hat{L}(f)\big| \leq B(\delta,n,\mathcal{F})
\label{eq:bartlett}
\end{align}
where $B(\delta,n,\mathcal{F})$ goes to zero as $n$ grows to infinity, that is the class $\mathcal{F}$ is learnable with respect to the loss~\cite{shalev2014understanding}.
Moreover $B(\delta,n,\mathcal{F})$ is usually an exponential bound which means that $B(\delta,n,\mathcal{F})$ grows logarithmically with respect to the inverse of $\delta$.

\begin{remark}
\label{rem:2}
If $\mathcal{F}$ is a compact subset of linear separators in a reproducing kernel Hilbert space, and the loss is Lipschitz in its first argument, then $B(\delta,n,\mathcal{F})$ can be obtained via Rademacher bounds~\cite{bartlett2002rademacher}.
In this case $B(\delta,n,\mathcal{F})$ goes to zero at least as ${\sqrt{1/n}}$ as $n$ grows and decreases with $\delta$ as ${\sqrt{\ln\left(1/\delta\right)}}$.
\end{remark}

We are now ready to state the first result of this section.
\begin{theorem}
\label{thm:mainresult1}
Let $\mathcal{F}$ be a learnable set of functions with respect to the loss function $\ell: \mathbb{R} \times {\cal Y} \rightarrow \mathbb{R}$, let $f^*$ be a solution of Problem (\ref{eq:alg:deterministic}) and let $\hat{f}$ be a solution of Problem (\ref{eq:alg:empirical}) with 
\begin{align}
\textstyle
\hat{\epsilon} = \epsilon + \sum_{k {=} 1}^{K} \sum_{q,q' {=} 1}^{Q} 
\sum_{p \in \{q,q'\}} 
B(\delta,n_{k,p},\mathcal{F}).
\nonumber
\end{align}
With probability at least $1- \delta$ it holds simultaneously that
\begin{align}
&
L(\hat{f}) - L(f^*) \leq 2
{\textstyle B\left(\frac{\delta}{(4 K Q^2 + 2)},n,\mathcal{F}\right)}, \nonumber \\
&  \sum_{k {=} 1}^{K} \sum_{p,q {=} 1}^{Q} \left| L^{k,p}_k(f) - L^{k,q}_k(f) \right| 
 \leq \epsilon + 2 \sum_{k {=} 1}^{K} \sum_{q,q' {=} 1}^{Q} 
\sum_{p \in \{q,q'\}} 
{\textstyle B\left(\frac{\delta}{(4 K Q^2 + 2)},n_{k,p},\mathcal{F}\right)}.
\nonumber
\end{align}
\end{theorem}
\begin{proof}
We first use Eq.~\eqref{eq:bartlett} to conclude that, with probability at least $1- 2 K Q^2 \delta$,
\begin{align}
& \textstyle \sup_{f \in \mathcal{F}} \left| 
\sum_{k {=} 1}^{K} \sum_{p,q {=} 1}^{Q} \big| L^{k,p}_k(f) {-} L^{k,q}_k(f) \big| {-} 
 \big| \hat{L}^{k,p}_k(f) {-} \hat{L}^{k,q}_k(f) \big|
\right| \nonumber \\
& 
\textstyle \leq 
\sum_{k {=} 1}^{K} \sum_{q,q' {=} 1}^{Q} 
\sum_{p \in \{q,q'\}} 
B(\delta,n_{k,p},\mathcal{F}).
\label{eq:proof1_eq2}
\end{align}
This inequality in turn implies that, with probability at least $1-2K Q^2 \delta$, it holds that
\begin{align}
&
\textstyle \Big\{ f: f \in \mathcal{F}, 
\sum_{k {=} 1}^{K} \sum_{p,q {=} 1}^{Q} \left| L^{k,p}_k(f) {-} L^{k,q}_k(f) \right|
\leq \epsilon \Big\} \label{eq:proof1_eq3} \\
&
\textstyle \subseteq \Big\{ f: f \in \mathcal{F}, 
\sum_{k {=} 1}^{K} \sum_{p,q {=} 1}^{Q} \left| \hat{L}^{k,p}_k(f) {-} \hat{L}^{k,q}_k(f) \right|
\leq \hat{\epsilon} \big\}.
\nonumber
\end{align}
Now, in order to prove the first statement of the theorem, let us decompose the excess risk as
\begin{align}
L(\hat{f}) {-} L(f^*\!) {=} 
L(\hat{f}) 
{-} \hat{L}(\hat{f})
{+} \hat{L}(\hat{f})
{-} \hat{L}(f^*\!) 
{+} \hat{L}(f^*\!)
{-} L(f^*\!).
\nonumber 
\end{align}
The inclusion property of Eq.~\eqref{eq:proof1_eq3} implies that $\hat{L}(\hat{f}) - \hat{L}(f^*) \leq 0$ with probability at least $1 -2 KQ^2 \delta$. Consequently with probability at least $1 - 2 KQ^2 \delta$ it holds that
\begin{align}
L(\hat{f}) - L(f^*)
\leq L(\hat{f}) - \hat{L}(\hat{f}) + 
\hat{L}(f^*) - L(f^*).
\nonumber 
\end{align}
The first statement now follows by Eq.~\eqref{eq:bartlett}.
As for the second statement, its proof consists in exploiting the results of Eqns.~\eqref{eq:proof1_eq2} and~\eqref{eq:proof1_eq3} together with a union bound.
\end{proof}
A consequence of the first statement of Theorem~\ref{thm:mainresult1} is that as $n$ tends to infinity $L(\hat{f})$ tends to a value which is not larger than $L(f^*)$, that is, G-FERM is consistent with respect to the risk of the selected model.
The second statement of Theorem~\ref{thm:mainresult1}, instead, implies that as $n$ tends to infinity we have that $\hat{f}$ tends to be $\epsilon$-fair.
In other words, G-FERM is consistent with respect to the fairness of the selected model.

\begin{remark}
Since $K,Q {\leq} n$ the bound in  Theorem~\ref{thm:mainresult1} behaves as $\sqrt{\ln\left(1/\delta\right)/n}$ in the same setting of Remark~\ref{rem:2} which is optimal~\cite{shalev2014understanding}.
\end{remark}

Thanks to Theorem~\ref{thm:mainresult1} we can state that $f^{*}$ is close to $\hat{f}$ both in term of its risk and its fairness.
Nevertheless, our final goal is to find an $f^*_h$ which solves the following problem 
\begin{align}\label{eq:problemHard}
\textstyle
\min_{f {\in} \mathcal{F}}\!\left\{\! {L}(f) {:} 
\sum_{k {=} 1}^{K} \sum_{p,q{=} 1}^{Q} \left| {P}^{k,p}(f) {-} {P}^{k,q}(f) \right| {\leq} \epsilon
\right\}.
\end{align}
Note that, the quantities in Problem~\eqref{eq:problemHard} cannot be computed since the underline data generating distribution is unknown.
Moreover, the objective function and the fairness constraint of Problem~\eqref{eq:problemHard} are non convex.

Theorem~\ref{thm:mainresult1} allow us to solve the first issue since we can safely search for a solution $\hat{f}_h$ of the empirical counterpart of Problem~\eqref{eq:problemHard}, which is given by
\begin{align}\label{eq:problemHardempirical}
\textstyle
\min_{f {\in} \mathcal{F}}\left\{\! \hat{L}(f) {:} 
\sum_{k {=} 1}^{K} \sum_{p,q {=} 1}^{Q} \left| \hat{P}^{k,p}(f) {-} \hat{P}^{k,q}(f) \right| {\leq} \hat{\epsilon}
\right\}
\end{align}
where 
\begin{align}
\textstyle
\hat{P}^{k,q}(f) {=}
\frac{1}{n_{k,q}} \sum_{(\boldsymbol{z},y) {\in} \mathcal{D}_{k,q}} \mathds{1} \left\{ f(\boldsymbol{z}) {\in} [t_k, t_{k{+}1}) \right\}.
\label{eq:phatlucapontil}
\end{align}
Unfortunately, Problem~\eqref{eq:problemHardempirical} is still a difficult non-convex non-smooth problem, and for this reason it is more convenient to solve a convex relaxation.
That is, we replace the possible non-convex loss function in the risk with its convex upper bound $\ell_c$ (e.g.~the square loss $\ell_{c} {=} (y{-}f(\boldsymbol{z}))^2$) and the losses $\ell_{k}$, $1{\leq}k{\leq}K$, in the constraint with a relaxation (e.g.~the linear loss $\ell_l({\hat y},y) {=} {\hat y} {-} y$) which allows to make the constraint convex.
In this way, we look for a solution $\hat{f}_c$ of the convex G-FERM problem
\begin{align}\label{eq:problemSoft}
\textstyle
\!\min_{f {\in} \mathcal{F}} \! \left\{\! \hat{L}_c(f) {:} 
\sum_{k {=} 1}^{K} \sum_{p, q {=} 1}^{Q} \! \! \left| \hat{L}_l^{k,p}(f) {-} \hat{L}_l^{k,q}(f) \right| \! {\leq} \hat{\epsilon} \!
\right\} \! .
\end{align}
Note that this approximation of the fairness constraint correspond to matching the first order moment~\cite{OnetoC060}.

The questions that arise here are whether $\hat{f}_c$ is close to $\hat{f}_h$, how much, and under which assumptions.
The following proposition sheds some lights on these issues.
\begin{proposition}
\label{thm:mainresult2}
If $\ell_c$ is a convex upper bound of the loss exploited to compute the risk then $
\hat{L}_{h}(f) \leq \hat{L}_{c}(f)$.
Moreover, if for $f: \mathcal{X} \rightarrow \mathbb{R}$ and for $\ell_l$
\begin{align*}
\textstyle
\sum_{k {=} 1}^{K} \sum_{p,q {=} 1}^{Q} \left| \hat{P}^{k,p}(f) {-} \hat{P}^{k,q}(f) \right| - \left| \hat{L}_l^{k,p}(f) {-} \hat{L}_l^{k,q}(f) \right| {\leq} \hat{\Delta}
\end{align*}
with $\hat{\Delta}$ small, then also the fairness is well approximated.
\end{proposition}

%Some comments are in order.
The first statement of Proposition~\ref{thm:mainresult2} tells us that exploiting the quality in approximating the risk depend on the quality of the convex approximation.
The second statement of Proposition~\ref{thm:mainresult2}, instead, tells us that if $\hat{\Delta}$ is small then the linear loss based fairness is close to the GF.
This condition is quite natural, empirically verifiable, and it has been exploited in previous work~\cite{maurer2004note,OnetoC060}.
Moreover, in Section~\ref{sec:5} we present experiments showing that $\hat{\Delta}$ is small. 

The bound in Proposition~\ref{thm:mainresult2} may be tighten by using different non-linear approximations of the GF.
However, the linear approximation proposed in this work gives a convex problem, and as we shall see in Section~\ref{sec:5}, works well in practice.

In summary, the combination of Theorem~\ref{thm:mainresult1} and Proposition~\ref{thm:mainresult2} provides conditions under which a solution $\hat{f}_c$ of Problem~\eqref{eq:alg:empirical}, which is convex, is close, {\em both in terms of risk and fairness measure}, to a solution $f^*_h$ of Problem~\eqref{eq:problemHard}, which is our final goal.
\section{G-FERM with Kernel Methods}
\label{sec:4}
In this section, we specify the G-FERM framework to the case that the underlying space of models is a reproducing kernel Hilbert space (RKHS)~\cite{shawe2004kernel,smola2001}.

We let $\kappa{:} \mathcal{Z} {\times} \mathcal{Z} {\rightarrow} \mathbb{R}$ be a positive definite kernel and let $\boldsymbol{\phi} {:} \mathcal{Z} {\rightarrow} \mathbb{H}$ be an induced feature mapping such that $\kappa(\boldsymbol{z},\boldsymbol{z}') {=} \langle \boldsymbol{\phi}(\boldsymbol{z}),\boldsymbol{\phi}(\boldsymbol{z}')\rangle$, for all $\boldsymbol{z},\boldsymbol{z}' {\in} \mathcal{Z}$, where $\mathbb{H}$ is the Hilbert space of square summable sequences.
Functions in the RKHS can be parametrized as 
\begin{equation}
f(\boldsymbol{z}) = \langle \boldsymbol{w} , \boldsymbol{\phi}(\boldsymbol{z})\rangle,~~~\boldsymbol{z} \in \mathcal{Z},
\label{eq:222}
\end{equation}
for some vector of parameters $\boldsymbol{w} \in \mathbb{H}$.
In practice a bias term (threshold) can be added to $f$ but to ease our presentation we do not include it here.

We propose to solve Problem~\eqref{eq:problemSoft} in the case that $\mathcal{F}$ is a ball in the RKHS and employ a convex loss function $\ell_{c}(y,{\hat y})$ to measure the empirical error.
Standard choices are the square loss in the case of regression or the hinge loss in the case of binary classification.
They are defined, for every $y,{\hat y} \in \mathbb{R}$, as $(y-{\hat y})^2$ and $\max(0,1-y{\hat y})$, respectively.
As for the fairness constraint we use the linear loss function $\ell_l$ which implies the constraint to be convex.
Then, we introduce the mean of the feature vectors associated with the training points restricted by the discretization of the sensitive feature and real outputs, namely
\begin{align}
\textstyle
\boldsymbol{u}_{k,q} = \frac{1}{N_{k,q}} \sum_{ (\boldsymbol{z},y) \in \mathcal{D}_{k,q}} \boldsymbol{\phi}(\boldsymbol{z}).
\end{align}
Using Eq.~\eqref{eq:222} the constraint in Problem~\eqref{eq:problemSoft} becomes
\begin{align}
\textstyle
\sum_{k {=} 1}^{K} \sum_{p,q {=} 1}^{Q} \left| \langle \boldsymbol{w},\boldsymbol{u}_{k,p} -\boldsymbol{u}_{k,q} \rangle \right| \leq \hat{\epsilon}
\end{align}
which can be written with more compact notation as $\|A^T w\|_1 {\leq} {\hat \epsilon}$, where $A$ is the matrix having as columns the vectors $\boldsymbol{u}_{k,p} {-} \boldsymbol{u}_{k,q}$.
With this notation, the fairness constraint can be interpret as the composition of ${\hat \epsilon}$ ball of the $\ell_1$ norm with a linear transformation $A$.

In practice, we solve the following Tikhonov regularization problem
\begin{align}
\textstyle
\min\limits_{\boldsymbol{w} \in \mathbb{H}} \quad
\sum_{(\boldsymbol{z},y) \in \mathcal{D}}
\ell_c(y,\langle \boldsymbol{w} , \boldsymbol{\phi}(\boldsymbol{z})\rangle) + \lambda \|\boldsymbol{w}\|^2 
, \quad
\text{s.t.} \quad\|A^\top w\|_1 \leq {\hat \epsilon},
\label{prob:ker}
\end{align}
where $\lambda$ is a positive parameter.
Note that, if $\hat{\epsilon} {=} 0$ the constraint reduces to the linear constraint $A^\top w {=}0$.

Problem~\eqref{prob:ker} can be kernelized by observing that, thanks to the Representer Theorem~\cite{shawe2004kernel}
\begin{equation}
\textstyle
\boldsymbol{w} \hspace{.1truecm}= \hspace{-.2truecm} \sum_{(\boldsymbol{z},y) \in \mathcal{D}} \boldsymbol{\phi}(\boldsymbol{z}).
\end{equation}

The dual of Problem~\eqref{prob:ker} may be derived using Fenchel duality, see e.g.~\cite[Theorem 3.3.5]{borwein2010convex}.
We postpone the discussion to future work since in our experiments we employed an off-the-shelf convex optimization solver\footnote{https://www.ibm.com/analytics/cplex-optimizer}.

Finally, we note that in the case when $\boldsymbol{\phi}$ is the identity mapping (i.e.~$\kappa$ is the linear kernel on $\mathbb{R}^d$) and $\hat{\epsilon}{=}0$ then the fairness constraint of Problem~\eqref{prob:ker} can be implicitly enforced by making a change of representation~\cite{OnetoC060}.
\section{Experiments}
\label{sec:5}
In this section we present a set of experiments to test the performance of the proposed method, both in terms of error and fairness.
We will study both the cases with categorical and continuous sensitive feature in the context of the regression (continuous label).
The classification task, as special case of our proposed framework, has been already studied in~\cite{OnetoC060}.
For this purpose, we selected two metrics to compare our method with the other baselines.
Concerning the error we collected the Mean Absolute Percentage Error (MAPE), that is equal to $\hat{L}(f)$ on the test set when $\ell(f(\boldsymbol{z}),y) = 100 \frac{|y - f(\boldsymbol{z})|}{|y|}$.
For what concerns the fairness of the model we will exploit the Differences of GF (DGF), see Definition~\ref{def:fairnessGF}, that is the following quantity, still estimated on the test set as
\begin{align*}
\textstyle
\text{DGF}(f) = 
\sum_{k {=} 1}^{K} \sum_{p,q {=} 1}^{Q} \!\! \left| \hat{P}^{k,p}(f) - \hat{P}^{k,q}(f) \right|
\end{align*}
where the expression of $\hat{P}^{k,p}(f)$ is given in Eq.~\eqref{eq:phatlucapontil}.

A set of four different algorithms is considered, with two different types of validation procedures.
The algorithms are divided in two groups: linear and non-linear kernels.
Concerning the linear methods, the baseline is regularized least squares (RLS), where we solve Problem~\eqref{prob:ker} with no fairness constraint and a linear kernel.
Fair RLS is our method in this category, that solves Problem~\eqref{prob:ker} with a linear kernel including the fairness constraint.
A kernel version of the same methods is KRLS, that solves Problem~\eqref{prob:ker} with no fairness constraint and a Gaussian kernel, i.e.~$\kappa(\boldsymbol{z},\boldsymbol{z}') = e^{-\gamma \| \boldsymbol{z} - \boldsymbol{z}' \|^2}$.
In comparison, our proposed algorithm is Fair KRLS, where we tackle Problem~\eqref{prob:ker} with the fairness constraint and a Gaussian kernel.

We follow two different types of possible validation procedures\footnote{Hyperparameters range: $\lambda$ ${\in}$ $\{10^{-4.0}, 10^{-3.5}, \cdots, 10^{+4.0} \}$ and $\gamma$ ${\in}$ $\{ 10^{-4},$ $10^{-3},$ $\cdots, $ $10^{+4} \}$.}.
The first one is standard, and we call it Naive Validation (Naive).
In particular, we performed a nested 10-fold cross validation (CV) to select the best hyperparameters and to test the final model.
This procedure is repeated 30 times, and we reported the average performance on the test set alongside its standard deviation.
A second validation procedure, called Novel Validation Procedure (NVP) as in~\cite{OnetoC060}, is slightly different and more focused on finding the best fair model among the ones with low error.
Also in this case, as general structure, we performed a nested 10-fold CV to test the final model.
For the inner part of the nested CV, we employ a two steps procedure.
In the first step, the 10-fold CV error for each of the combination of the hyperparameters is computed.
In the second step, we shortlist all the hyperparameters' combinations with error close to the best one (in our case, above 90\% of the best MAPE).
Finally, from this list, we select the hyperparameters with the lowest DGF.

For the sake of completeness, all the experiments have been performed both having and not having the sensitive feature in the model's functional form, i.e.~the sensitive feature is available (or not available) at test time.
\subsection{Datasets}
For the purpose of testing the proposed proposed methodology we employed two different datasets for regression.

The first one is a classic benchmark dataset for fairness called Communities and Crime dataset\footnote{http://archive.ics.uci.edu/ml/datasets/communities+and+crime} (CRIME).
CRIME combines socioeconomic data and crime rate data on communities in the United States.
In the case of categorical sensitive feature, following~\cite{calders2013controlling}, we made a binary attribute $s$ as to the percentage of black population, which yielded $970$ instances of $s {=} 1$ with a mean crime rate $0.35$ and $1024$ instances of $s {=} 0$ with a mean crime rate $0.13$.
In this case $\mathcal{S}_Q {=} \{-0.5,0.5,1.5\}$.
Concerning the experiments with continuous sensitive feature we maintain the real value of the percentage of black population, avoiding the binarization step of it and then we consider $Q {=} 5$ and a uniform set $\mathcal{S}_Q$ over $[0,1]$, i.e.~$\mathcal{S}_Q {=} \{0.0,0.2,\dots,0.8,1.0\}$.

The second dataset we propose is new and it has been collected at the University of Genoa (UNIGE).
This dataset is a proprietary and highly sensitive dataset containing all the data about the past and present students enrolled at the UNIGE.
In this study we take into consideration students who enrolled, in the academic year (a.y.) 2017-2018.
The dataset contains $5000$ instances, each one described by $35$ attributes (both numeric and categorical) about ethnicity, gender, financial status, and previous school experience.
The scope is to predict the average grades and the end of the first semester.
In the case of categorical sensitive feature, we consider as sensitive feature the gender ($s {=} 1$ female and $s {=} 0$ male) and consequently $\mathcal{S}_Q {=} \{-0.5,0.5,1.5\}$.
In the context of continuous sensitive attribute, we select as sensitive feature the income of the student, with $Q {=} 5$ following the official separation in five bins from the tuition system of the University of Genoa (details at link \url{https://www.studenti.unige.it/tasse/importi/}).

\begin{table}%[t]
\centering
%\small
\setlength{\tabcolsep}{0.1cm}
\renewcommand{\arraystretch}{1.1}
\begin{tabular}{|l|c|c|c|c|c|c|c|c|c|c|c|c|c|c|c|c|c|c|c|c|}
\hline
\hline
& \multicolumn{2}{c|}{CRIME} 
& \multicolumn{2}{c|}{UNIGE} \\
Method & MAPE & DGF & MAPE & DGF \\
\hline
\hline 
\multicolumn{5}{c}{Sensitive Feature not included in the model's functional form.}\\
\hline
\hline
Naive RLS & $9.1{\pm}0.5$ & $0.19{\pm}0.06$ & $21.2{\pm}1.8$ & $0.29{\pm}0.08$ \\
NVM RLS & $10.2{\pm}0.8$ & $0.16{\pm}0.05$ & $23.4{\pm}1.9$ & $0.23{\pm}0.09$ \\
NVM Fair RLS & $10.5{\pm}1.0$ & $0.11{\pm}0.04$ & $24.2{\pm}1.9$ & $0.15{\pm}0.09$ \\
\hline
Naive KRLS & $8.7{\pm}0.4$ & $0.18{\pm}0.05$ & $12.2{\pm}0.8$ & $0.19{\pm}0.05$ \\
NVM KRLS & $8.9{\pm}0.7$ & $0.17{\pm}0.05$ & $13.7{\pm}1.1$ & $0.12{\pm}0.05$ \\
NVM Fair KRLS & $9.0{\pm}0.7$ & $0.11{\pm}0.04$ & $14.1{\pm}1.2$ & $0.06{\pm}0.03$ \\
\hline
\hline 
\multicolumn{5}{c}{Sensitive Feature included in the model's functional form.}\\
\hline
\hline
Naive RLS & $9.1{\pm}0.6$ & $0.20{\pm}0.05$ & $19.7{\pm}1.7$ & $0.33{\pm}0.11$ \\
NVM RLS & $9.5{\pm}0.6$ & $0.18{\pm}0.05$ & $21.9{\pm}1.9$ & $0.28{\pm}0.09$ \\
NVM Fair RLS & $9.5{\pm}0.7$ & $0.12{\pm}0.03$ & $21.8{\pm}1.8$ & $0.19{\pm}0.10$ \\
\hline
Naive KRLS & $8.5{\pm}0.6$ & $0.19{\pm}0.04$ & $11.5{\pm}0.8$ & $0.21{\pm}0.06$ \\
NVM KRLS & $8.6{\pm}0.6$ & $0.18{\pm}0.05$ & $12.6{\pm}0.9$ & $0.13{\pm}0.05$ \\
NVM Fair KRLS & $8.7{\pm}0.7$ & $0.12{\pm}0.04$ & $12.9{\pm}0.9$ & $0.07{\pm}0.03$ \\
\hline
\hline
\end{tabular}
\caption{Results with $\hat{\epsilon} = 0$ and $K = 10$.}
\label{tab:results1}
\end{table}

\begin{figure}
\centering
\includegraphics[width=.495\columnwidth]{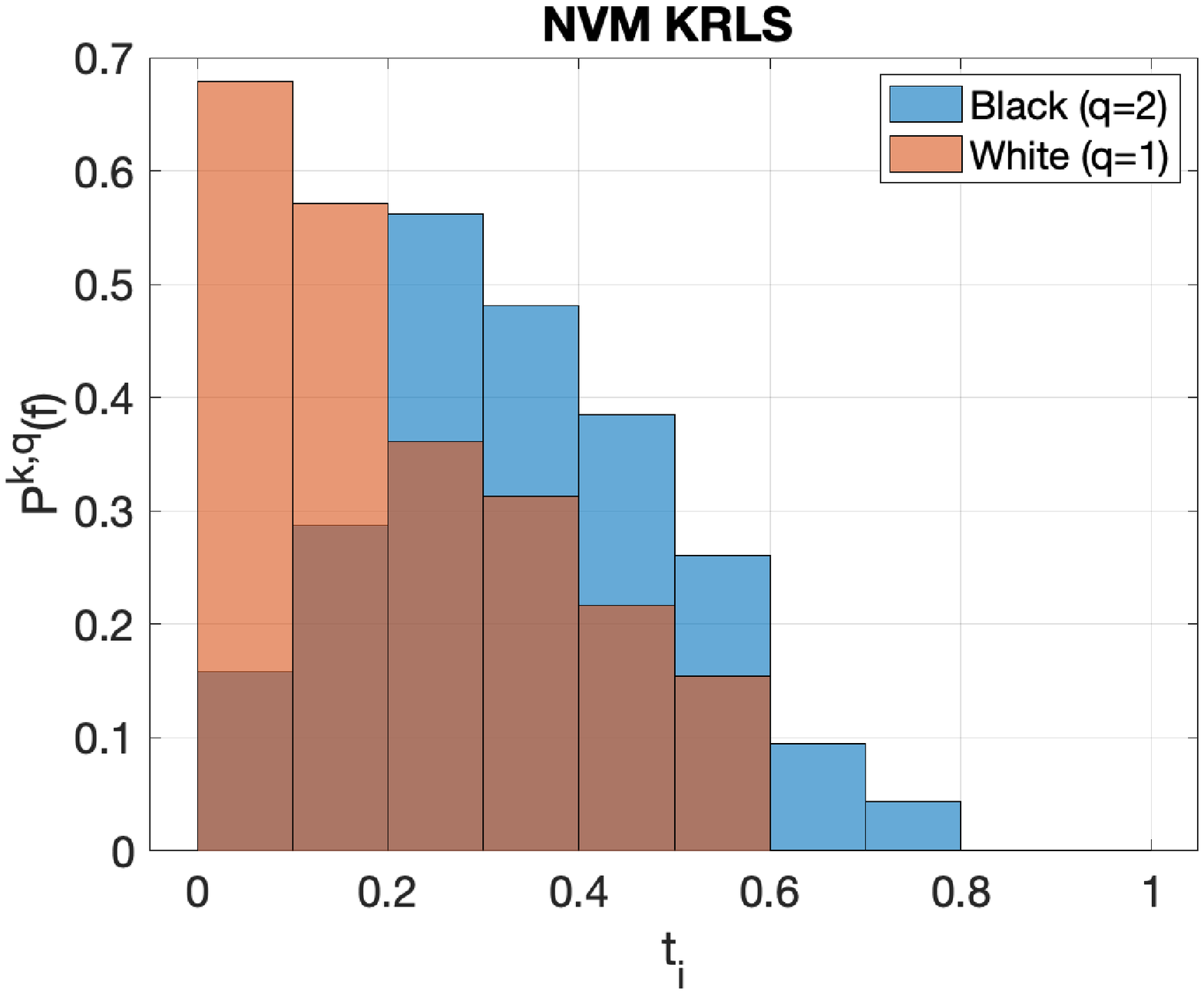} 
\includegraphics[width=.495\columnwidth]{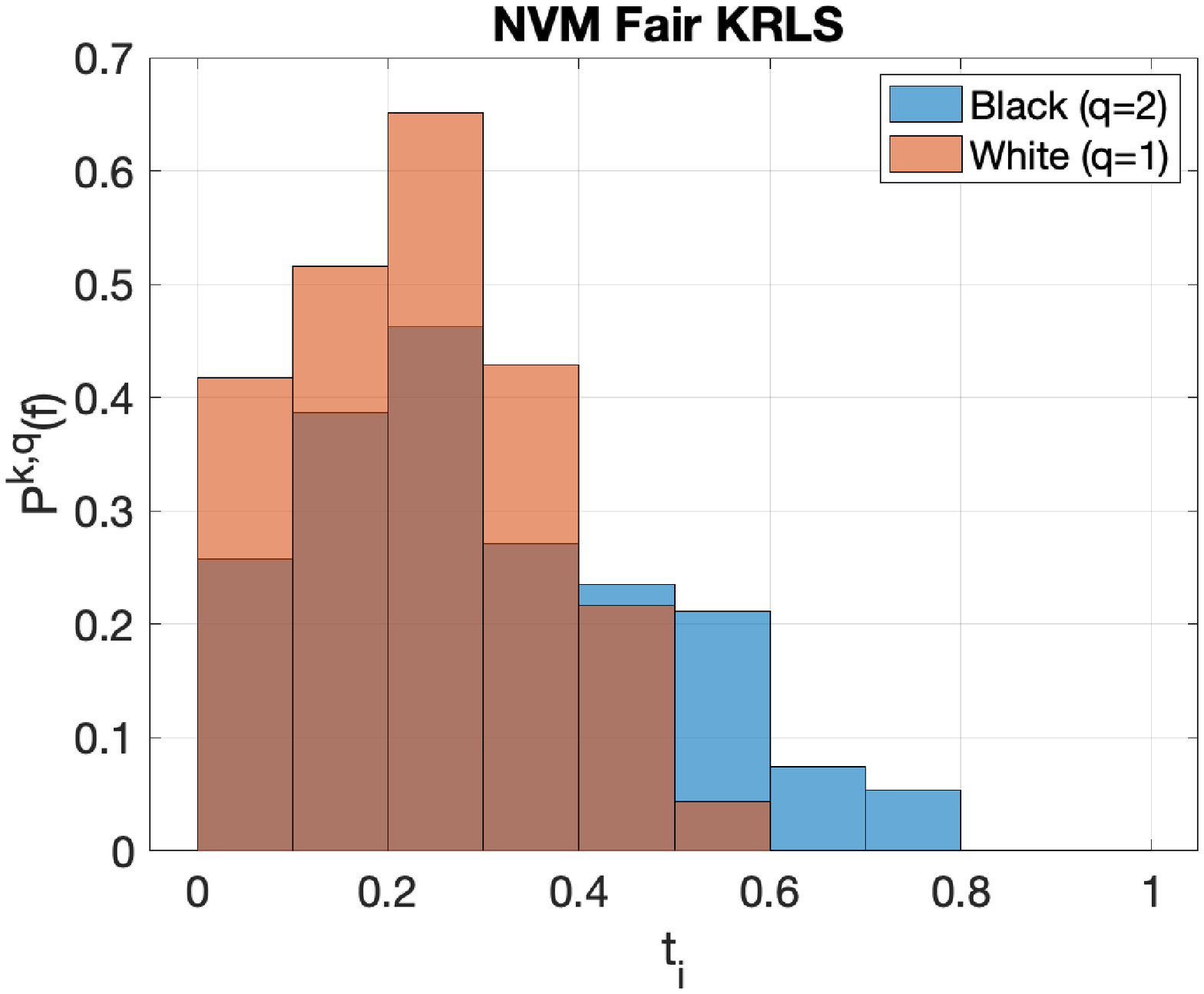}
\caption{Two overlapped (White $q{=}1$ Black $q{=}2$) histograms of $\mathbb{P}^{k,q}$ for the CRIME dataset with NVM KRLS and NVM Fair KRLS when the sensitive feature not included in the function form of the model.}
\label{fig:histog}
\end{figure}

\begin{table}%[t]
%\small
\centering
\setlength{\tabcolsep}{0.1cm}
\renewcommand{\arraystretch}{1.1}
\begin{tabular}{|l|c|c|c|c|c|c|c|c|c|c|c|c|c|c|c|c|c|c|c|c|}
\hline
\hline
& CRIME 
& UNIGE \\
Method & $\hat{\Delta}$ & $\hat{\Delta}$ \\
\hline
\hline 
\multicolumn{3}{c}{Sensitive Feature not included} \\
\multicolumn{3}{c}{in the model's functional form.}\\
\hline
\hline
NVM Fair RLS & $0.03$ & $0.02$ \\
\hline
NVM Fair KRLS & $0.03$ & $0.03$ \\
\hline
\hline 
\multicolumn{3}{c}{Sensitive Feature included} \\
\multicolumn{3}{c}{in the model's functional form.}\\
\hline
\hline
NVM Fair RLS & $0.04$ & $0.03$ \\
\hline
NVM Fair KRLS & $0.03$ & $0.03$ \\
\hline
\hline
\end{tabular}
\caption{$\hat{\Delta}$ with $\hat{\epsilon} = 0$ and $K = 10$.}
\label{tab:delta}
%\vspace{-.5cm}
\end{table}

\iffalse
\begin{table*}
\scriptsize
\centering
\setlength{\tabcolsep}{0.1cm}
\renewcommand{\arraystretch}{1.1}
\begin{tabular}{|l|c|c|c|c|c|c|c|c|c|c|c|c|c|c|c|c|c|c|c|c|}
\hline
\hline
Dataset
& \multicolumn{6}{c|}{CRIME}
& \multicolumn{6}{c|}{UNIGE}\\
& \multicolumn{2}{c|}{$\hat{\epsilon} =$ 0}
& \multicolumn{2}{c|}{$\hat{\epsilon} =$ 0.005}
& \multicolumn{2}{c|}{$\hat{\epsilon} =$ 0.01}
& \multicolumn{2}{c|}{$\hat{\epsilon} =$ 0}
& \multicolumn{2}{c|}{$\hat{\epsilon} =$ 0.005}
& \multicolumn{2}{c|}{$\hat{\epsilon} =$ 0.01} \\
Method & MAPE & DGF & MAPE & DGF & MAPE & DGF & MAPE & DGF & MAPE & DGF & MAPE & DGF \\
\hline
\hline 
\multicolumn{13}{c}{Sensitive Feature not included in the model's functional form.}\\
\hline
\hline
NVM Fair RLS & $10.5$ & $0.11$ & $10.3$ & $0.14$ & $10.2$ & $0.16$ & $24.2$ & $0.15$ & $23.7$ & $0.19$ & $23.4$ & $0.23$ \\
\hline
NVM Fair KRLS & $9.0$ & $0.11$ & $8.9$ & $0.14$ & $8.9$ & $0.17$ & $14.1$ & $0.06$ & $13.9$ & $0.09$ & $13.7$ & $0.12$ \\
\hline
\hline 
\multicolumn{13}{c}{Sensitive Feature included in the model's functional form.}\\
\hline
\hline
NVM Fair RLS & $9.5$ & $0.12$ & $9.5$ & $0.16$ & $9.5$ & $0.18$ & $21.8$ & $0.19$ & $21.8$ & $0.24$ & $21.9$ & $0.28$ \\
\hline
NVM Fair KRLS & $8.7$ & $0.12$ & $8.6$ & $0.17$ & $8.6$ & $0.18$ & $12.9$ & $0.07$ & $12.7$ & $0.09$ & $12.6$ & $0.13$ \\
\hline
\hline
\end{tabular}
\caption{Results varying $\hat{\epsilon}$ with $K = 10$}
\label{tab:results2}
\end{table*}
\fi

\begin{table}%[t]
%\small
\centering
\setlength{\tabcolsep}{0.1cm}
\renewcommand{\arraystretch}{1.1}
\begin{tabular}{|l|c|c|c|c|c|c|c|c|c|c|c|c|c|c|c|c|c|c|c|c|}
\hline
\hline
& \multicolumn{2}{c|}{$\hat{\epsilon} =$ 0}
& \multicolumn{2}{c|}{$\hat{\epsilon} =$ 0.005}
& \multicolumn{2}{c|}{$\hat{\epsilon} =$ 0.01} \\
Method & MAPE & DGF & MAPE & DGF & MAPE & DGF  \\
\hline
\hline 
\multicolumn{7}{c}{CRIME}\\
\hline
\hline 
\multicolumn{7}{c}{Sensitive Feature not included in the model's functional form.}\\
\hline
\hline
NVM Fair RLS  & $10.5$ & $0.11$ & $10.3$ & $0.14$ & $10.2$ & $0.16$ \\
\hline
NVM Fair KRLS &  $9.0$ & $0.11$ & $8.9$  & $0.14$ &  $8.9$ & $0.17$ \\
\hline
\hline 
\multicolumn{7}{c}{Sensitive Feature included in the model's functional form.}\\
\hline
\hline
NVM Fair RLS & $9.5$ & $0.12$ & $9.5$ & $0.16$ & $9.5$ & $0.18$ \\
\hline
NVM Fair KRLS & $8.7$ & $0.12$ & $8.6$ & $0.17$ & $8.6$ & $0.18$ \\
\hline
\hline
\multicolumn{7}{c}{UNIGE} \\
\hline
\hline 
\multicolumn{7}{c}{Sensitive Feature not included in the model's functional form.}\\
\hline
\hline
NVM Fair RLS   & $24.2$ & $0.15$ & $23.7$ & $0.19$ & $23.4$ & $0.23$ \\
\hline
NVM Fair KRLS & $14.1$ & $0.06$ & $13.9$ & $0.09$ & $13.7$ & $0.12$ \\
\hline
\hline 
\multicolumn{7}{c}{Sensitive Feature included in the model's functional form.}\\
\hline
\hline
NVM Fair RLS & $21.8$ & $0.19$ & $21.8$ & $0.24$ & $21.9$ & $0.28$ \\
\hline
NVM Fair KRLS & $12.9$ & $0.07$ & $12.7$ & $0.09$ & $12.6$ & $0.13$ \\
\hline
\hline
\end{tabular}
\caption{Results varying $\hat{\epsilon}$ with $K = 10$}
\label{tab:results2}
%\vspace{-.5cm}
\end{table}

\iffalse
\begin{table*}
\scriptsize
\centering
\setlength{\tabcolsep}{0.1cm}
\renewcommand{\arraystretch}{1.1}
\begin{tabular}{|l|c|c|c|c|c|c|c|c|c|c|c|c|c|c|c|c|c|c|c|c|}
\hline
\hline
Dataset
& \multicolumn{6}{c|}{CRIME}
& \multicolumn{6}{c|}{UNIGE}\\
& \multicolumn{2}{c|}{$K =$ 5} 
& \multicolumn{2}{c|}{$K =$ 10}
& \multicolumn{2}{c|}{$K =$ 20}
& \multicolumn{2}{c|}{$K =$ 5} 
& \multicolumn{2}{c|}{$K =$ 10}
& \multicolumn{2}{c|}{$K =$ 20} \\
Method & MAPE & DGF & MAPE & DGF & MAPE & DGF & MAPE & DGF & MAPE & DGF & MAPE & DGF \\
\hline
\hline 
\multicolumn{13}{c}{Sensitive Feature not included in the model's functional form.}\\
\hline
\hline
NVM Fair RLS & $10.4$ & $0.13$ & $10.5$ & $0.11$ & $15.5$ & $0.05$ & $23.6$ & $.019$ & $24.2$ & $0.15$ & $35.7$ & $0.06$ \\
\hline
NVM Fair KRLS & $9.0$ & $0.14$ & $9.0$ & $0.11$ & $14.8$ & $0.04$ & $13.7$ & $.010$ & $14.1$ & $0.06$ & $22.4$ & $0.03$ \\
\hline
\hline 
\multicolumn{13}{c}{Sensitive Feature included in the model's functional form.}\\
\hline
\hline
NVM Fair RLS & $9.5$ & $0.16$ & $9.5$ & $0.12$ & $13.8$ & $0.05$ & $21.8$ & $0.25$ & $21.8$ & $0.19$ & $33.9$ & $0.09$ \\
\hline
NVM Fair KRLS & $8.7$ & $0.15$ & $8.7$ & $0.12$ & $13.7$ & $0.04$ & $12.8$ & $0.11$ & $12.9$ & $0.07$ & $21.8$ & $0.03$ \\
\hline
\hline
\end{tabular}
\caption{Results varying $K$ with $\hat{\epsilon} = 0$}
\label{tab:results3}
%\vspace{-.5cm}
\end{table*}
\fi

\begin{table}%[t]
%\small
\centering
\setlength{\tabcolsep}{0.1cm}
\renewcommand{\arraystretch}{1.1}
\begin{tabular}{|l|c|c|c|c|c|c|c|c|c|c|c|c|c|c|c|c|c|c|c|c|}
\hline
\hline
Dataset
& \multicolumn{2}{c|}{$K =$ 5} 
& \multicolumn{2}{c|}{$K =$ 10}
& \multicolumn{2}{c|}{$K =$ 20}\\
Method & MAPE & DGF & MAPE & DGF & MAPE & DGF \\
\hline
\hline 
\multicolumn{7}{c}{CRIME}\\
\hline
\hline 
\multicolumn{7}{c}{Sensitive Feature not included in the model's functional form.}\\
\hline
\hline
NVM Fair RLS & $10.4$ & $0.13$ & $10.5$ & $0.11$ & $15.5$ & $0.05$ \\
\hline
NVM Fair KRLS & $9.0$ & $0.14$ & $9.0$ & $0.11$ & $14.8$ & $0.04$ \\
\hline
\hline 
\multicolumn{7}{c}{Sensitive Feature included in the model's functional form.}\\
\hline
\hline
NVM Fair RLS & $9.5$ & $0.16$ & $9.5$ & $0.12$ & $13.8$ & $0.05$ \\
\hline
NVM Fair KRLS & $8.7$ & $0.15$ & $8.7$ & $0.12$ & $13.7$ & $0.04$ \\
\hline
\hline 
\multicolumn{7}{c}{UNIGE}\\
\hline
\hline 
\multicolumn{7}{c}{Sensitive Feature not included in the model's functional form.}\\
\hline
\hline
NVM Fair RLS   & $23.6$ & $.019$ & $24.2$ & $0.15$ & $35.7$ & $0.06$ \\
\hline
NVM Fair KRLS & $13.7$ & $.010$ & $14.1$ & $0.06$ & $22.4$ & $0.03$ \\
\hline
\hline 
\multicolumn{7}{c}{Sensitive Feature included in the model's functional form.}\\
\hline
\hline
NVM Fair RLS & $21.8$ & $0.25$ & $21.8$ & $0.19$ & $33.9$ & $0.09$ \\
\hline
NVM Fair KRLS & $12.8$ & $0.11$ & $12.9$ & $0.07$ & $21.8$ & $0.03$ \\
\hline
\hline
\end{tabular}
\caption{Results varying $K$ with $\hat{\epsilon} = 0$}
\label{tab:results3}
%\vspace{-.5cm}
\end{table}
\subsection{Results and Discussion}
Results for regression tasks with categorical sensitive feature are presented in Table~\ref{tab:results1}, where MAPE and DGF are shown for the different datasets (CRIME and UNIGE), algorithms (RLS and KRLS), validation procedure (Naive and NVM), with and without the fairness constraints, and availability of the sensitive feature at test time.

For both datasets, it is clear the advantage of using our method in order to obtain more fair models (i.e.~lower DGF) at the expenses of a slightly higher error (i.e.~higher MAPE).
Moreover, having the sensitive feature at test time increases model accuracy (i.e.~lower MAPE) and reduces the fairness measure (i.e.~higher DGF).
The improvement is stronger in the kernel case, and where the original unfairness of the standard method is higher.

An important question concerns the sensitivity of our method with respect to the parameter $\hat{\epsilon}$ (acceptable unfairness) and the number of bins $K$.
Tables~\ref{tab:results2} and~\ref{tab:results3} reports this analysis.
We repeated the same experimental procedure of Table~\ref{tab:results1} for both datasets (CRIME and UNIGE), and algorithms (RLS and KRLS), and possible availability of the sensitive feature at test time, when the fairness constraint is active and with the NVM.
We let $\hat{\epsilon}$ range in $\{ 0, 0.005, 0.001 \}$ with fixed $K{=}10$, and also let $K$ range in $\{5, 10, 20 \}$ maintaining $\hat{\epsilon} {=} 0$.
The results confirm our theoretical insights. Making $\hat{\epsilon}$ larger induces lower MAPE and larger DGF, confirming the trade-off between error and fairness.
Considering $K$, we have that larger values of $K$ corresponds to impose a higher number of constraints, something that impacts negatively the MAPE value (i.e.~the higher $K$, the higher MAPE).
On the other hand, increasing the value of $K$ makes the final model more fair, with a lower DGF.

Figure~\ref{fig:histog} shows the different behaviours of the standard non-linear regression models (without fairness constraints, i.e.~NVM KRLS) and our method (NVM Fair KRLS) over the CRIME dataset, specifically when the sensitive feature is not part of the model's functional form.
In particular, we reported the different element in the summation which composes the DGF: $P^{k,q}(f)$ for White ($q{=}1$) and Black ($q{=}2$).
Our method, (right plot) obtains two probability distributions among the two different groups that are more similar with respect the baseline (left plot).
This suggests that our method is more fair with respect to the selected sensitive feature.

We collected in Table~\ref{tab:delta} the $\hat{\Delta}$ values (see Proposition~\ref{thm:mainresult2}), for both datasets, for both NVM Fair RLS and NVM Fair KRLS, with and without the sensitive feature in the model's functional form.
As it can be noted, the value $\hat{\Delta}$ remains small and, consequently, our method provides a good convex approximation of the original non-convex optimization problem of Eq.~\eqref{eq:problemHardempirical} in practice.

As a final experiment, we empirically demonstrate that it is possible to generate fair models with continuous sensitive features.
Table~\ref{tab:results1bis} reports the results for NVM KRLS and NVM Fair KRLS for both datasets with and without the sensitive feature in the functional form of the model.
The obtained MAPE and DGF confirm the results described above in the case of categorical sensitive attributes, empirically demonstrating that our methodology is able to tackle the regression tasks having categorical and continuous sensitive feature.
\begin{table}
\centering
%\small
\setlength{\tabcolsep}{0.1cm}
\renewcommand{\arraystretch}{1.1}
\begin{tabular}{|l|c|c|c|c|c|c|c|c|c|c|c|c|c|c|c|c|c|c|c|c|}
\hline
\hline
& \multicolumn{2}{c|}{CRIME} 
& \multicolumn{2}{c|}{UNIGE} \\
Method & MAPE & DGF & MAPE & DGF \\
\hline
\hline 
\multicolumn{5}{c}{Sensitive Feature not included in the model's functional form.}\\
\hline
\hline
NVM KRLS 	& $ 8.9{\pm}0.7$ & $0.17{\pm}0.05$ & $15.9{\pm}1.3$ & $0.16{\pm}0.06$ \\
NVM Fair KRLS 	& $10.5{\pm}0.8$ & $0.05{\pm}0.02$ & $17.8{\pm}1.4$ & $0.04{\pm}0.02$ \\
\hline
\hline 
\multicolumn{5}{c}{Sensitive Feature included in the model's functional form.}\\
\hline
\hline
NVM KRLS 	& $ 8.6{\pm}0.6$ & $0.18{\pm}0.05$ & $14.5{\pm}1.3$ & $0.19{\pm}0.07$ \\
NVM Fair KRLS 	& $10.1{\pm}0.8$ & $0.06{\pm}0.03$ & $16.2{\pm}1.4$ & $0.05{\pm}0.02$ \\
\hline
\hline
\end{tabular}
\caption{Results with $\hat{\epsilon} = 0$, $K = 10$ and $Q = 5$.}
\label{tab:results1bis}
%\vspace{-.5cm}
\end{table}
\section{Conclusion and Future Work}
\label{sec:6}
In this work, we studied the problem of enhancing supervised learning with fairness requirements.
We presented a framework based on empirical risk minimization under a novel and generalized fairness constraint.
Contrarily to the previous methods, our approach can handle both regression and classification problems and both continuous or categorical sensitive attributes.
Furthermore we observed that our approach generalizes and reduces to known approaches available in literature.
We addressed the statistical properties of the method and considered a convex relaxation of the fairness constraint, which can be linked to the non-convex constraint by means of a data dependent bound.
We instantiated this approach in the setting of kernel methods, for which the convex fairness constraint can be efficiently implemented both implicitly and explicitly.
Finally, we provided experimental results on two real-world datasets that indicate the effectiveness of our approach in comparison with some baselines which either do not impose the fairness constraint or impose the constraint during the validation procedure.
Future work will be devoted to extend the range of applicability of our method and to study tighter bounds under specialized conditions.
\bibliographystyle{plain}
\bibliography{biblio}
\end{document}